\PassOptionsToPackage{table}{xcolor}

\documentclass[10pt,twocolumn,letterpaper]{article}

\usepackage[pagenumbers]{cvpr} 
\usepackage{multirow}
\usepackage{comment}

\usepackage[dvipsnames]{xcolor}

\definecolor{cvprblue}{rgb}{0.21,0.49,0.74}
\usepackage[pagebackref,breaklinks,colorlinks,citecolor=cvprblue]{hyperref}
\usepackage[table]{xcolor}
\usepackage{bm}
\usepackage{threeparttable}
\usepackage{multirow}
\usepackage{graphicx}
\usepackage{multicol}
\usepackage{comment}

\usepackage{amsmath}
\usepackage{amssymb}
\usepackage{cuted}
\usepackage{gensymb}
\usepackage{tablefootnote}
\usepackage{listings}

\definecolor{codegreen}{rgb}{0,0.6,0}
\definecolor{codegray}{rgb}{0.5,0.5,0.5}
\definecolor{codepurple}{rgb}{0.58,0,0.82}
\definecolor{backcolour}{rgb}{0.95,0.95,0.92}
\lstdefinestyle{mystyle}{
    backgroundcolor=\color{backcolour},   
    commentstyle=\color{codegreen},
    keywordstyle=\color{magenta},
    stringstyle=\color{codepurple},
    basicstyle=\ttfamily\footnotesize,
    breakatwhitespace=false,         
    breaklines=true,                 
    captionpos=b,                    
    keepspaces=true,                 
    showspaces=false,                
    showstringspaces=false,
    showtabs=false,                  
    tabsize=2
}
\title{Compression of 3D Gaussian Splatting with Optimized \\Feature Planes and Standard Video Codecs}

\author{
Soonbin Lee
\and
Fangwen Shu
\and
Yago Sánchez
\and
Thomas Schierl
\and
Cornelius Hellge
\and
Fraunhofer Heinrich-Hertz-Institute (HHI), Germany\\
{\tt\small \{first\_name\}.\{last\_name\}@hhi.fraunhofer.de}
}

\begin{document}

\maketitle
\begin{abstract}
3D Gaussian Splatting is a recognized method for 3D scene representation, known for its high rendering quality and speed. However, its substantial data requirements present challenges for practical applications. In this paper, we introduce an efficient compression technique that significantly reduces storage overhead by using compact representation. We propose a unified architecture that combines point cloud data and feature planes through a progressive tri-plane structure. Our method utilizes 2D feature planes, enabling continuous spatial representation. To further optimize these representations, we incorporate entropy modeling in the frequency domain, specifically designed for standard video codecs. We also propose channel-wise bit allocation to achieve a better trade-off between bitrate consumption and feature plane representation. Consequently, our model effectively leverages spatial correlations within the feature planes to enhance rate-distortion performance using standard, non-differentiable video codecs. Experimental results demonstrate that our method outperforms existing methods in data compactness while maintaining high rendering quality. Our project page is available at \href{https://fraunhoferhhi.github.io/CodecGS}{https://fraunhoferhhi.github.io/CodecGS}. 
\end{abstract}    
\section{Introduction}
\label{sec:intro}

Novel view synthesis aims to generate new perspectives of a 3D scene or object by interpolating from a limited set of images with known camera parameters. Recent advancements, such as Neural Radiance Fields (NeRF) \cite{mildenhall2021nerf, barron2021mip, barron2022mip}, employ implicit neural representations to learn a radiance field from which novel views can be synthesized. However, the computational cost of neural network evaluations significantly impacts both the efficiency of training and rendering. Recently, 3D Gaussian Splatting (3DGS) \cite{kerbl20233d} has emerged as a method for generating volumetric scene representations using gaussian primitives, which can be rendered at high speed on GPUs. This approach represents the scene as a collection of 3D points, each characterized by learnable gaussian attributes. These parameters are optimized through differentiable rasterization to accurately represent a set of input images. However, representing 3D scenes often requires millions of gaussians, consuming several gigabytes of storage and memory. This presents a significant challenge for rendering on devices with limited computing resources, such as mobile devices or head-mounted displays. 

Recent studies have made early progress in compressing 3DGS by reducing both the number and size of 3DGS \cite{lee2023compact, fan2023lightgaussian, niedermayr2023compressed}. These methods typically employ point pruning to remove gaussians with minimal impact on rendering quality, followed by vector quantization to convert the remaining continuous attributes into a discrete set of codewords for further compression. However, these approaches often result in suboptimal performance, as they mainly focus on reducing rendering distortion while neglecting redundancies within gaussian primitives. As a result, these limitations hinder the development of more compact 3D scene representations.

\begin{figure}[t]
 \centering 
 \includegraphics[width=\linewidth]{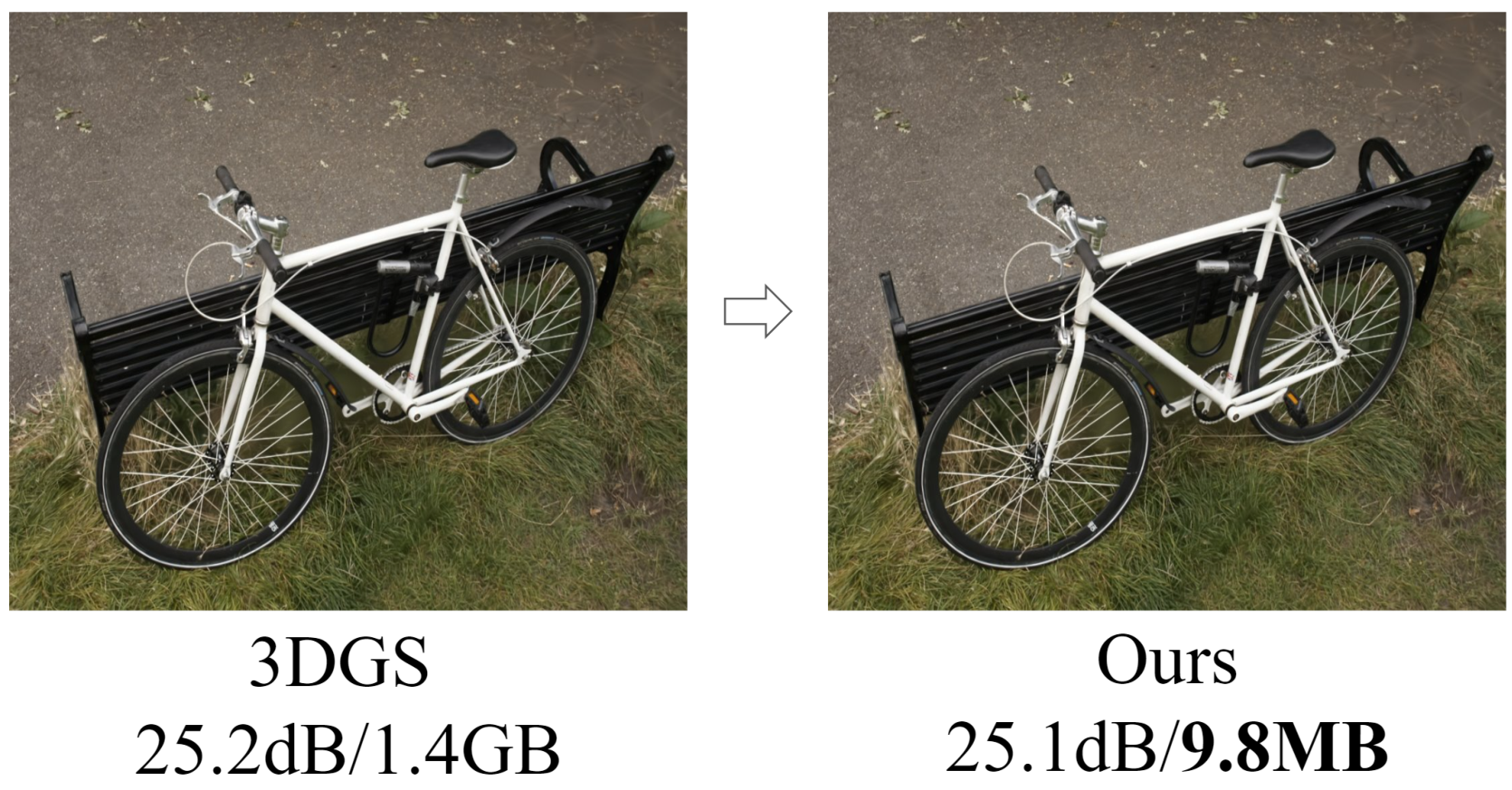}
 \caption{Our method achieves a \textbf{146$\times$} compression with negligible loss in image quality, a significant improvement over 3DGS \cite{kerbl20233d} the `bicycle' scene. Our method seamlessly integrates with standard video codecs and utilizes the original 3DGS rendering pipeline, achieving comparable rendering speeds with minimal overhead.}
 \label{fig:fig1}
\end{figure}

We address the storage challenge of 3DGS in a different way. Drawing inspiration from previous work \cite{videorf10655718,tetrirf10656750}, we develop a compression method that significantly reduces storage requirements by incorporating standard video codecs. To achieve this, we propose a feature plane architecture that implicitly predicts all gaussian primitives. These primitives are rendered using the original 3DGS pipeline, which includes a differentiable gaussian rasterizer. Building on these plane representations, we incorporate an entropy modeling that facilitates learning a more refined representation. Our key contributions are summarized as follows:

\begin{itemize}
\item We present a compact architecture for representing the attributes of 3DGS using feature planes. For training, we propose point initialization and progressive training techniques, which ensure stable training and enable seamless integration with existing 3DGS pipelines.
\item We propose a frequency-domain entropy parameterization for the feature plane, in which the original data is transformed using a block-wise discrete cosine transform (DCT). Experiments show that this parameterization significantly enhances compression performance for a standard, non-differentiable codec.
\item To achieve better rate-distortion performance, we propose entropy weighting techniques based on channel importance scores. We also incorporate techniques for feature plane representation, including channel-to-frame concatenation and piecewise projective contraction.
\end{itemize}

The experimental results show that our method significantly reduces storage requirements compared to 3DGS. For experimental data sets, our approach achieves high quality rendering with only a few MB of storage (typically $<$10MB) without significant loss in rendering quality. This highlights the effectiveness of our method in optimizing storage efficiency and rendering quality.


\section{Related Work}
\label{sec:relatedwork}

\begin{figure*}[t]
 \centering 
 \includegraphics[width=\linewidth]{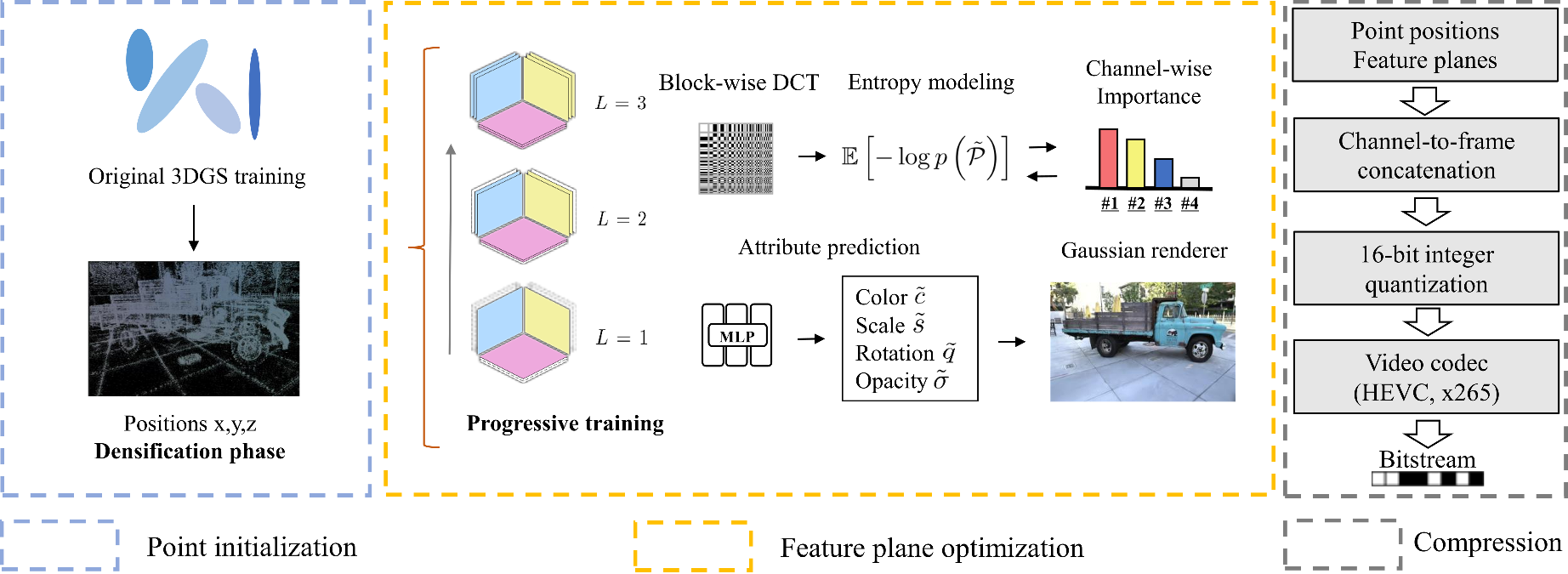}
 \caption{\textbf{Overview of the proposed model.} Following the original 3DGS densification, we train the model to predict all gaussian attributes using the feature plane architecture. The feature plane achieves a more compact representation through our proposed DCT entropy modeling and channel bit allocation techniques, which leads to performance improvements.}
 \label{fig:fig2}
\end{figure*}

\textbf{Radiance Field Compression.} NeRF presents an implicit approach to 3D scene representation that achieves high visual quality, though its substantial computational requirements pose challenges for real-time rendering \cite{mildenhall2021nerf}. Subsequent methods \cite{muller2022instant, chen2022tensorf, sun2022direct} have aimed to optimize training speed by incorporating explicit grid structures and other data-efficient frameworks. However, these explicit structures often add significant storage overhead. To address this, various techniques have been developed to reduce grid complexity, including vector quantization \cite{takikawa2022variable}, wavelet transforms applied to grid-based neural fields \cite{rho2023masked}, parameter pruning \cite{li2023compressing}, context-aware entropy encoding \cite{shin2023binary, far10655258} and Fourier-based transformations \cite{ReRF, lee2023ecrfentropyconstrainedneuralradiance}. 

Recent advancements, such as VideoRF \cite{videorf10655718} and TeTriRF \cite{tetrirf10656750}, leverage tensor factorization and standard video codecs to achieve further compression gains. However, these methods primarily target structured grids and are not directly applicable to 3DGS, which relies on unstructured gaussian primitives lacking consistent spatial organization. Inspired by this related work, we propose an effective framework for 3DGS compression.

\textbf{Gaussian Splatting Compression.} 3DGS is designed as an efficient scene representation that employs primitives to achieve fast high-quality rendering. Despite these benefits, the high storage requirements of 3DGS make it challenging to deploy in resource-constrained environments. Research efforts have focused on compressing 3DGS by reducing the number and complexity of gaussian points. For example, compression methods based on entropy modeling \cite{girish2023eagles, wang2024rdogaussian}, sensitivity-based pruning \cite{niedermayr2023compressed, reduce10.1145/3651282, fan2023lightgaussian}, and learnable masking approaches \cite{lee2023compact} aim to minimize the impact of non-essential gaussians, effectively reducing data size. Notably, Self-Organizing Gaussian \cite{morgenstern2024compact3dscenerepresentation} applies sorting algorithms to compress attributes using image codecs. The other approach uses Scaffold-GS \cite{lu2024scaffold}, which utilizes the anchor points and their relationships \cite{hac2024, compgs10.1145/3664647.3681468}. Although these approaches achieve significant data reductions, they often overlook potential redundancies in the gaussian attributes. As a result, these methods may overlook opportunities for even more efficient compression without compromising visual quality.

\textbf{Video Coding.} For video coding, High Efficiency Video Coding (HEVC) \cite{sullivan2012overview} and Versatile Video Coding (VVC) \cite{bross2021overview} are renowned for their effective rate-distortion optimization, significantly compressing 2D video data by exploiting spatial correlations. Their success has inspired attempts to integrate these codecs into neural radiance field models, harnessing their ability to reduce redundancy within structured 2D grids \cite{videorf10655718, tetrirf10656750}. However, handling 3DGS data with these technologies remains unexplored due to the lack of organized spatial structure.


\section{Proposed Method}
\subsection{Feature Plane Representations with 3DGS}
3DGS uses gaussian primitives to generate volumetric data for 3D space rendering. Each gaussian primitive consists of attributes including position $\boldsymbol{\mu}$, spherical harmonics (SH) color $\boldsymbol{c}$, scale $\boldsymbol{s}$, rotation $\boldsymbol{q}$, and opacity $\boldsymbol{\sigma}$. Since these parameters can number in the millions depending on the scene, explicitly storing all attributes requires significant storage overhead. We adopt the $k$-planes model \cite{fridovich2023k} for efficient attribute representation. This model derives compact features by factorizing over planes using the Hadamard product, then decodes these features through a small MLP $g$. We employ Tri-plane, a static version of Hex-plane without the time axis. Tri-plane $\mathcal{P}$ takes 3D point positions $\mathbf{x}$ as input and predicts the corresponding attributes for each point.

\begin{equation}
\begin{aligned}
g(\mathcal{P}(\mathbf{x})) = \{\tilde{c}, \tilde{s}, \tilde{q}, \tilde{\sigma}\}
\end{aligned}
\end{equation}


Despite its high expressiveness, learning attributes through feature planes from scratch presents significant challenges. Training high-resolution, multi-channel planes can be slow to converge, which impacts point densification. We observed that directly replacing the feature plane to predict all attributes causes inconsistencies with point densifications. This conflict between plane and point optimization leads to suboptimal results. 

To address this, we employ a two-phase approach. First, we run 15k iterations using the original 3DGS training until the point densification phase concludes. Then, we begin feature plane training to predict all gaussian attributes, including color, scale, rotation, and opacity. This method allows us to maintain a constant number of points from the start of feature plane training, while continuously refining their positions.

\subsection{Progressive Training}
However, training with feature planes remains unstable, as 3DGS points represent sparse signals that may not be suitable for interpolation. To address this, we introduce progressive training as a key strategy. Recent work \cite{Heo10.5555/3618408.3618936, kim2024synergistic} has proposed curriculum learning for grid models. These studies have shown promising results for various vision tasks by dynamically adjusting the learning rate for each grid level. Similarly, we introduce a progressive channel masking strategy. This strategy selectively updates parameters only for the currently activated channel levels, preventing backpropagation for inactive channels. During the $T_i$ iteration stage, backpropagation is limited to the $[0,L_{i}]$ channel of feature planes. The value of $L$ increases with each stage, enabling progressive training of feature planes.

This strategy enables stable training of feature planes, leading to better reconstruction quality. In the early stages, the feature planes capture coarse geometric information with limited capacity. As training progresses, the feature plane gradually incorporates finer levels, adding more detail to the coarse structure. This progressive training results in a multi-level representation, where lower levels capture global and low-frequency information, while higher levels capture high-frequency details.

Building on this observation, we propose a channel importance-based bit allocation strategy. This method calculates the sensitivity of each channel level and allocates bits accordingly, enhancing overall performance. A detailed explanation of this method is provided in Section \ref{sec:sec3.4}.

\subsection{Feature Plane Optimization}
\textbf{How can we optimize the feature plane to achieve more efficient compression of video codecs?} Video codec technologies enable hardware decoding on a wide range of devices, offering competitive performance with low complexity. These features make them promising candidates for integration with 3DGS compression techniques. However, conventional standard video codecs are non-differentiable, which prevents the establishment of an explicit optimization loss. A common approach is to use the $\mathcal{L}_1$ loss to sparsify the signal, but this may lead to significant information loss.

We revisit the entropy modeling approach, which aims to minimize the entropy of quantized parameters \cite{balle2016end, balle2018variational, balle2020nonlinear}. Specifically, we can introduce an entropy parameterization technique to the feature plane $\mathcal{P}$. Based on this approach, the model minimizes its entropy $I$ with uniform noise $\mathcal{U}$ to use straight through estimator (STE). Although recent studies have proposed entropy modeling for 3DGS compression \cite{hac2024, compgs10.1145/3664647.3681468, wang2024rdogaussian}, these studies did not take video codecs into consideration.

\begin{equation}
\begin{aligned}
I(\mathcal{P}) =& \mathbb{E} \left[ -\log p \left( \tilde{\mathcal{P}} \right) \right], 
\\ 
 \tilde{\mathcal{P}} = \mathcal{P} + u, &  u \sim \mathcal{U} \left( -\frac{1}{Q_{\text{step}}}, \frac{1}{Q_{\text{step}}} \right)
\end{aligned}
\end{equation}

This loss makes the parameter more robust to quantization errors, but we observed that it does not yield better results with video codec compression. The main reason is that standard codec apply DCT for signal compression in the frequency domain \cite{sullivan2012overview, bross2021overview}. In our method, unorganized 3D points are projected onto a tri-plane, enabling us to apply an $N$$\times$$M$ block-wise DCT $\mathcal{F}$ to the $(x,y)$ coordinates of each plane. This approach treats the planes as conventional image signals, effectively exploiting their spatial correlation.

\begin{equation}
\begin{aligned}
\mathcal{F}(\mathcal{P})_{u,v} = \sum_{x=0}^{N-1} \sum_{y=0}^{M-1} \mathcal{P}_{x, y} \quad \quad \quad
\\
\cos \left( \frac{\pi (2x + 1) u}{2N} \right) \cos \left( \frac{\pi (2y + 1) v}{2M} \right)
\end{aligned}
\end{equation}

\begin{figure}[t]
 \centering
      \begin{subfigure}[b]{0.49\linewidth}
         \centering
         \includegraphics[width=0.98\linewidth]{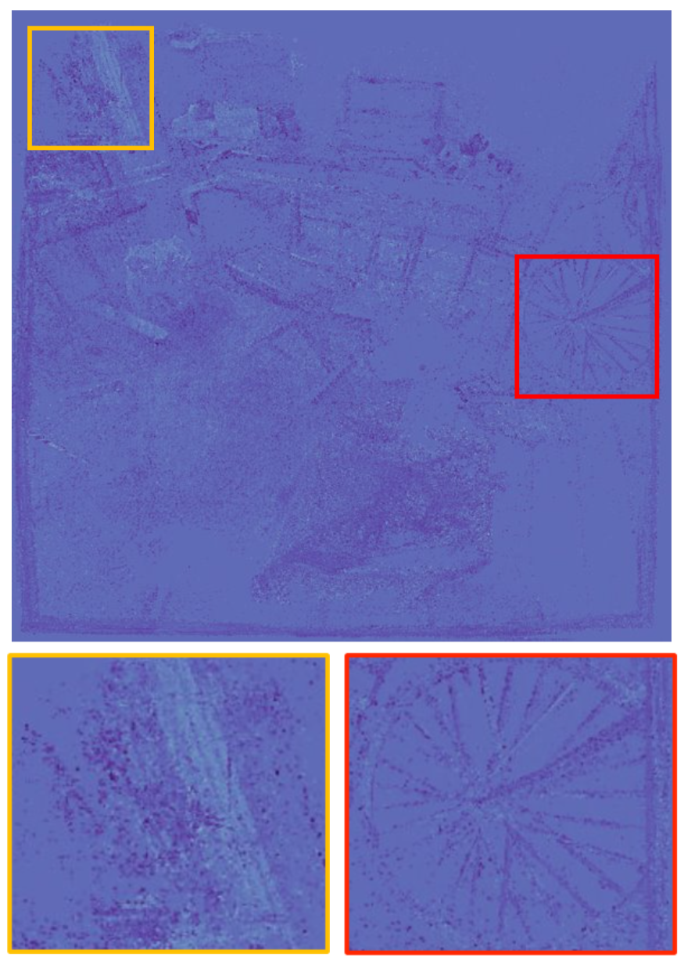}
         \caption{$\mathcal{P}_{1}^{XY}$ with $\mathcal{L}_{1}$}
     \end{subfigure}
     \vspace{0.01\linewidth}
          \begin{subfigure}[b]{0.49\linewidth}
         \centering
         \includegraphics[width=0.98\linewidth]{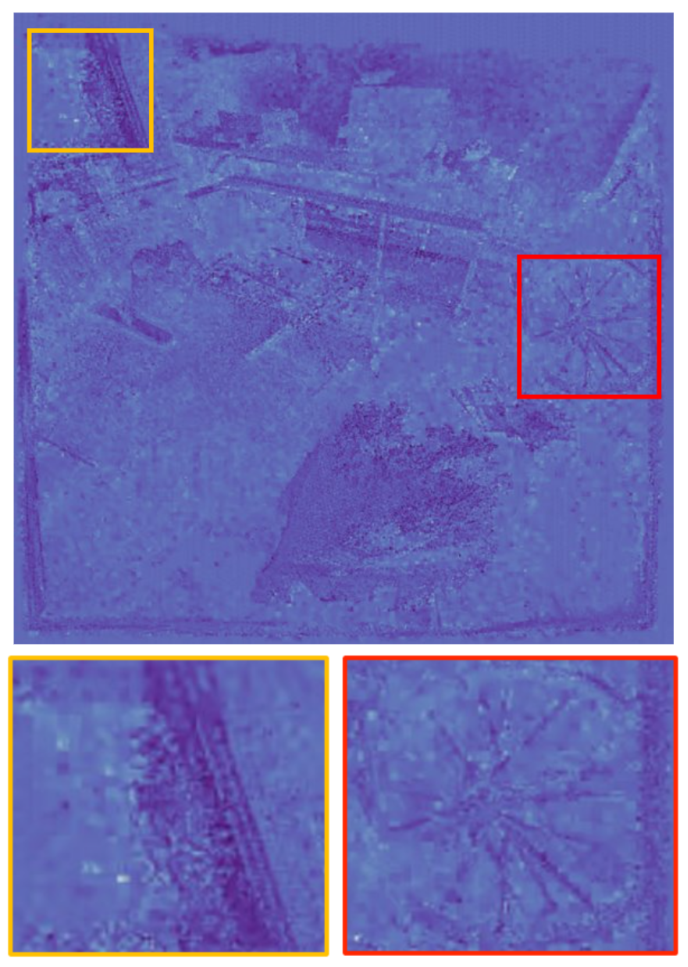}
         \caption{$\mathcal{P}_{1}^{XY}$ with $\mathcal{L}_{ent}$}
     \end{subfigure}
 \caption{\textbf{Visualization of the first channel in the XY plane $\mathcal{P}_{1}^{XY}$ for the \textit{`Bonsai'} scene.} For better visualization, we used a very high term for each loss. (a) with sparsity $\mathcal{L}_{1}$ loss and (b) with proposed DCT entropy $\mathcal{L}_{ent}$ loss. We use the piecewise-projective contraction detailed in Section \ref{sec:sec4.1}.
 }
 \label{fig:fig3}
\end{figure}

That is, the entropy optimization target is not the plane parameters, but their transformed coefficients. We found the minimization of the $I(\mathcal{F}(\mathcal{P}))$ is highly beneficial for video codec performance. Note that our method uses a 2D DCT for each channel of planes with a 4$\times$4 block size, which is commonly used as the minimum transform unit (TU) size. After training, feature planes are normalized to the range $[0,1]$ and then scaled to 16-bit integers, corresponding to the YUV 16-bit format. Unlike \cite{compgs10.1145/3664647.3681468}, which introduced a learnable quantization step size $Q_{step}$, video codec may not be correctly incorporated into this optimization process. Therefore, we must empirically determine the $Q_{step}$ for transformed coefficients. Assuming with the 16-bit scalar quantization mentioned above, we found that a $Q_{step}$ of $2^{8}$ yields the best results. A detailed analysis of various $Q_{step}$ values is available in the appendix. Our proposed method jointly optimizes the entropy of the transformed coefficients, resulting in a more compact representation.

Fig. \ref{fig:fig3} illustrates how the optimized plane preserves signal through block-wise approximation, in contrast to sparsity $\mathcal{L}_{1}$ loss. This method more effectively maintains the original signal, resulting in superior compression performance. Our experimental results show that this representation improves rate-distortion performance for standard codec without requiring any modifications to the codec software.

\begin{table*}[]
\resizebox{\textwidth}{!}{
\begin{tabular}{l|cccc|cccc|cccc}
\hline
                          & \multicolumn{4}{c|}{Mip-NeRF360~\cite{barron2022mip}}                                                                                             & \multicolumn{4}{c|}{DeepBlending~\cite{hedman2018deep}}                                                                                            & \multicolumn{4}{c}{Tank \& Temples~\cite{knapitsch2017tanks}}                                                                                           \\ \cline{2-13} 
\multirow{-2}{*}{Methods} & \multicolumn{1}{c}{PSNR$\uparrow$}      & \multicolumn{1}{c}{SSIM$\uparrow$}      & \multicolumn{1}{c}{LPIPS$\downarrow$}     & \multicolumn{1}{c|}{Size$\downarrow$}    & \multicolumn{1}{c}{PSNR$\uparrow$}      & \multicolumn{1}{c}{SSIM$\uparrow$}      & \multicolumn{1}{c}{LPIPS$\downarrow$}     & \multicolumn{1}{c|}{Size$\downarrow$}    & PSNR$\uparrow$                          & SSIM$\uparrow$                          & LPIPS$\downarrow$                         & Size$\downarrow$                         \\ \hline
3DGS~\cite{kerbl20233d}                      & 27.49                         & 0.813 & 0.222 & 745                          & 29.42                         & 0.899                         & 0.247                         & 664                          & 23.69                         & 0.844                         & 0.178 & 431.0                        \\
Scaffold-GS~\cite{lu2024scaffold}                & 27.50                         & 0.806                         & 0.252                         & 254                          & 30.21                         & 0.906                        & 0.254                         & 66.0                         & 23.96                         & 0.853 & 0.177 & 86.50                        \\ \hline
Self-Organizing~\cite{morgenstern2024compact3dscenerepresentation}            & 26.01                         & 0.772                         & 0.259                         & 23.9                         & 28.92                         & 0.891                         & 0.276                         & \colorbox[HTML]{FFDAC0}{8.40}                         & 22.78                         & 0.817                         & 0.211                         & 13.1                         \\
EAGLES~\cite{girish2023eagles}                    & 27.15                         & \colorbox[HTML]{FFDAC0}{0.808}                         & \colorbox[HTML]{FFFFC7}{0.238}                         & 68.9                         & \colorbox[HTML]{FFDAC0}{29.91}                         & \colorbox[HTML]{FFCCC9}{0.910} & \colorbox[HTML]{FFCCC9}{0.250}                         & 62.0                         & 23.41                         & 0.840                         & 0.200                         & 34.0                         \\
LightGaussian~\cite{fan2023lightgaussian}             & 27.00                         & 0.799                         & 0.249                         & 44.5                         & 27.01                         & 0.872                         & 0.308                         & 33.9                         & 22.83                         & 0.822                         & 0.242                         & 22.4                         \\
Compact3DGS~\cite{lee2023compact}                & 27.08                         & 0.798                         & 0.247                         & 48.8                         & 29.79                         & 0.901                         & 0.258                         & 43.2                         & 23.32                         & 0.831                         & 0.201                         & 39.4                         \\
C3DGS~\cite{niedermayr2023compressed}         & 26.98                         & 0.801                         & 0.238                         & 28.8                         & 29.38                         & 0.898                         & 0.253 & 25.3                         & 23.32                         & 0.832                         & \colorbox[HTML]{FFFFC7}{0.194}                      & 17.3                         \\
RDOGaussian~\cite{wang2024rdogaussian}           & 27.05                         & 0.802                         & 0.239 & 23.4                         & 29.63                         & 0.902                         & \colorbox[HTML]{FFFFC7}{0.252}                         & 18.0                         & 23.34                         & 0.835                         &  0.195                         & 12.0                         \\ 
CompGS~\cite{compgs10.1145/3664647.3681468}               & \colorbox[HTML]{FFFFC7}{27.26}                         & 0.802                         & 0.239                         & \colorbox[HTML]{FFFFC7}{16.5}                         & 29.69                         & 0.900                         & 0.280                         & 8.77                         & \colorbox[HTML]{FFCCC9}{23.71}                          & \colorbox[HTML]{FFFFC7}{0.840}                          & 0.210                         & \colorbox[HTML]{FFFFC7}{9.61}                         \\ 
HAC~\cite{hac2024}                   & \colorbox[HTML]{FFCCC9}{27.53}                         & \colorbox[HTML]{FFFFC7}{0.807}                         & \colorbox[HTML]{FFDAC0}{0.238}                         & \colorbox[HTML]{FFDAC0}{15.3}                         & \colorbox[HTML]{FFCCC9}{30.19}                         & \colorbox[HTML]{FFFFC7}{0.905}                         & 0.262                         & \colorbox[HTML]{FFCCC9}{7.46} & \colorbox[HTML]{FFDAC0}{23.70}                         & \colorbox[HTML]{FFCCC9}{0.846}                         & \colorbox[HTML]{FFCCC9}{0.185}                         & \colorbox[HTML]{FFDAC0}{8.44} \\ \hline
Ours                  & \colorbox[HTML]{FFDAC0}{27.30}                         & \colorbox[HTML]{FFCCC9}{0.810}                         & \colorbox[HTML]{FFCCC9}{0.236}                         & \colorbox[HTML]{FFCCC9}{9.78} & \colorbox[HTML]{FFFFC7}{29.82} & \colorbox[HTML]{FFDAC0}{0.907}                         & \colorbox[HTML]{FFDAC0}{0.251}                         & \colorbox[HTML]{FFFFC7}{8.62} & \colorbox[HTML]{FFFFC7}{23.63}                         & \colorbox[HTML]{FFDAC0}{0.842}                         & \colorbox[HTML]{FFDAC0}{0.192}                         & \colorbox[HTML]{FFCCC9}{7.46}                         \\ \hline
\bottomrule
\end{tabular} }
\caption{\textbf{Quantitative results for the compression of 3DGS.} We conducted experiments using the Mip-NeRF360, DeepBlending, and Tank \& Temples datasets. The results are highlighted in \colorbox[HTML]{FFCCC9}{red}, \colorbox[HTML]{FFDAC0}{orange} and \colorbox[HTML]{FFFFC7}{yellow}, respectively. We have highlighted only the compression models, and the comparative metrics are sourced from the respective papers.}
\label{tab:tabs1}
\end{table*}

\begin{figure}[t]
 \centering
      \begin{subfigure}[b]{0.32\linewidth}
         \centering
         \includegraphics[width=0.98\linewidth]{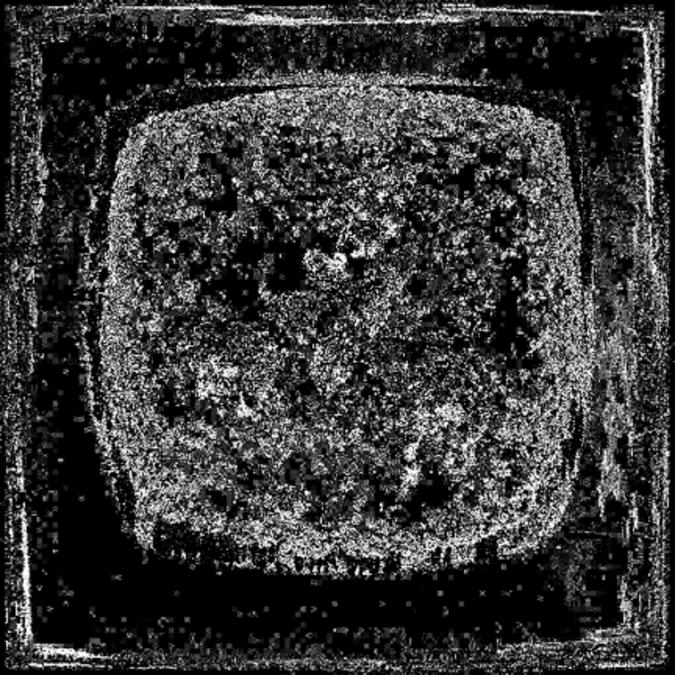}
         \caption{$\mathcal{P}_{1}^{XZ}$}
     \end{subfigure}
     \vspace{0.01\linewidth}
          \begin{subfigure}[b]{0.32\linewidth}
         \centering
         \includegraphics[width=0.98\linewidth]{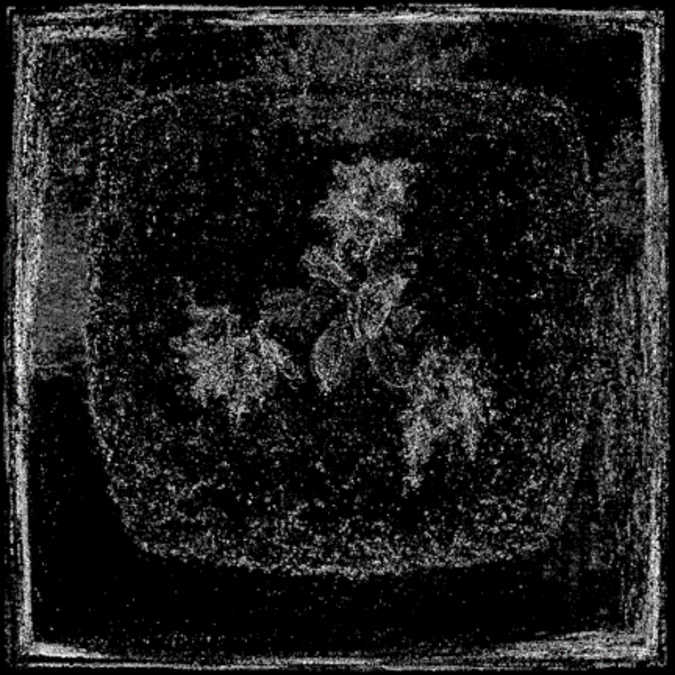}
         \caption{$\mathcal{P}_{3}^{XZ}$}
     \end{subfigure}
     \vspace{0.01\linewidth}
          \begin{subfigure}[b]{0.32\linewidth}
         \centering
         \includegraphics[width=0.98\linewidth]{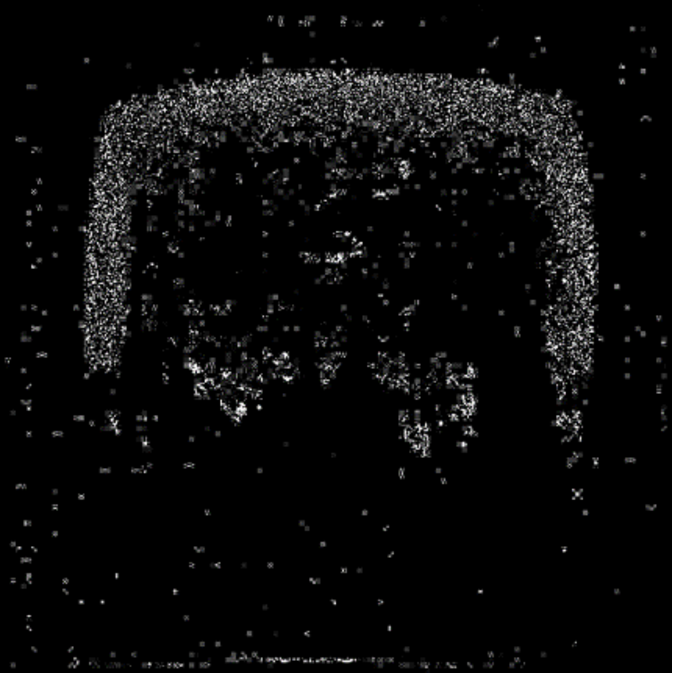}
         \caption{$\mathcal{P}_{5}^{XZ}$}
     \end{subfigure}
 \caption{\textbf{Visualization of the 1st, 3rd, and 5th channels of XZ plane with progressive training for the \textit{`Flowers'} scene.} The iteration stages $T_{i}$ are \{0, 5000, 10000\} with corresponding $L_{i}$ values of \{2, 4, 6\} for 30k iteration training. With the progressive training, parameter energy is mainly concentrated in the lower-level channels, while the higher-level channels show sparser representations. 
 }
 \label{fig:fig4}
\end{figure}

\subsection{Channel Importance-based Bit Allocation}
\label{sec:sec3.4}
With progressive training, each channel affects visual quality differently. Fig. \ref{fig:fig4} shows the impact of each channel using this strategy. Based on this observation, we establish criteria to optimize the rate-distortion trade-off across channels. This paper introduces channel importance (CI) as a method to determine bit allocation for each channel. Specifically, we derive the CI scores by calculating the sensitivity \cite{niedermayr2023compressed}, but applied to the feature plane $\mathcal{P}$ rather than to the gaussian parameters. This sensitivity can be represented as the gradient of each channel $\mathcal{P}_{c}$ with respect to the gradient of energy $E$, which corresponds to the RGB values of all training views, with the number of pixels denoted by $P$. Therefore, a large CI magnitude indicates high sensitivity to changes in visual quality.

\begin{equation}
CI_{c}(\mathcal{P})=\frac{1}{\sum_{i=1}^{train} P_i} \left|\frac{\partial E_i}{\partial \mathcal{P}_{c}}\right|
\end{equation}

Using this score, we can design a straightforward bit allocation method, such as assigning higher quantization parameters (QP) with the video codec to less significant channels. Alternatively, we propose a more effective approach by using CI to determine the weight factor $w$ for the $c$-th channel in the DCT entropy loss. 

\begin{equation}
w_{c}=\frac{CI_{1}(\mathcal{P})}{CI_{c}(\mathcal{P})}
\end{equation}

For simplicity, we set $w_{1}$ to 1 for the first channel, while higher-level $w_{i,i>1}$ typically have values greater than 1. A higher value of $w_{c}$ further reduces entropy, resulting in a lower bitrate allocation for channel $c$. This weighting factor $w_{c}$ allocates bitrate for each channel, based on its importance. In other words, this method automatically determines $w_{c}$ to achieve better rate-distortion optimization. Experimental results show that further improvement can be achieved while avoiding exhaustive hyperparameter search.


\begin{equation}
\mathcal{L}_{ent}=\sum_{c \in C}^{C}w_{c}I(\mathcal{F}(\mathcal{P}_{c}))
\end{equation}

In addition, we apply a small amount of $\mathcal{L}_{1}$ loss to reduce unnecessary areas in the randomly initialized feature planes, which contributes to size reduction.

\begin{equation}
\mathcal{L}_{\text {1}}=\|\mathcal{P}\|_{1}
\end{equation}

We combine the 3DGS rendering loss $\mathcal{L}_{\text{render}}$ with our proposed loss, resulting in a total supervision given by:
\begin{equation}
\mathcal{L}=\mathcal{L}_{render}+\lambda_{ent}\mathcal{L}_{ent}+\lambda_{1}\mathcal{L}_{1}
\end{equation}

The feature planes incorporate this loss function to integrate conventional video codec. By adjusting the parameter $\lambda_{ent}$ in various experiments, the feature plane effectively balances the trade-off between compressed size and reconstructed visual quality.

\begin{figure*}[t]
 \centering
      \begin{subfigure}[b]{0.15\linewidth}
         \centering
         \includegraphics[width=0.98\linewidth]{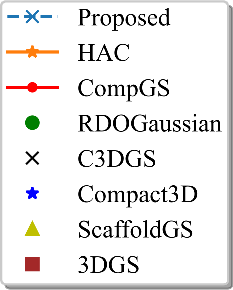}
     \end{subfigure}
          \begin{subfigure}[b]{0.27\linewidth}
         \centering
         \includegraphics[width=0.98\linewidth]{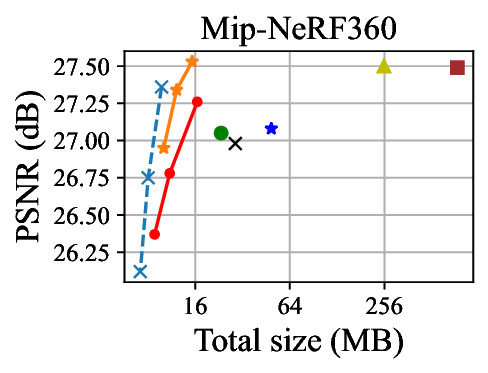}
     \end{subfigure}
          \begin{subfigure}[b]{0.27\linewidth}
         \centering
         \includegraphics[width=0.98\linewidth]{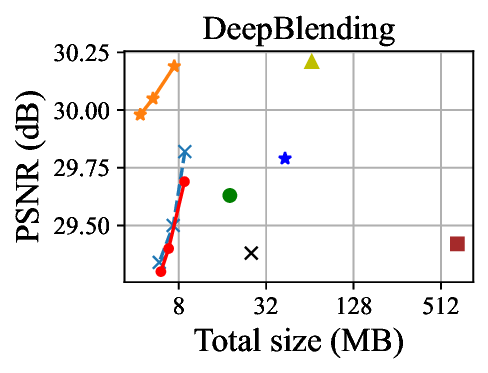}
     \end{subfigure}
          \begin{subfigure}[b]{0.27\linewidth}
         \centering
         \includegraphics[width=0.98\linewidth]{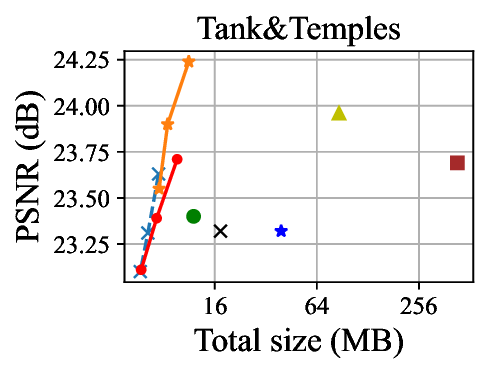}
     \end{subfigure}
 \caption{\textbf{RD curves for quantitative comparisons.} Rate-distortion (RD) plots are provided for each dataset using 3DGS compression models. For the x-axis, a log$_{2}$ scale is used for better visualization.
 }
 \label{fig:fig5}
\end{figure*}

\begin{figure*}[t]
 \centering 
 \includegraphics[width=\linewidth]{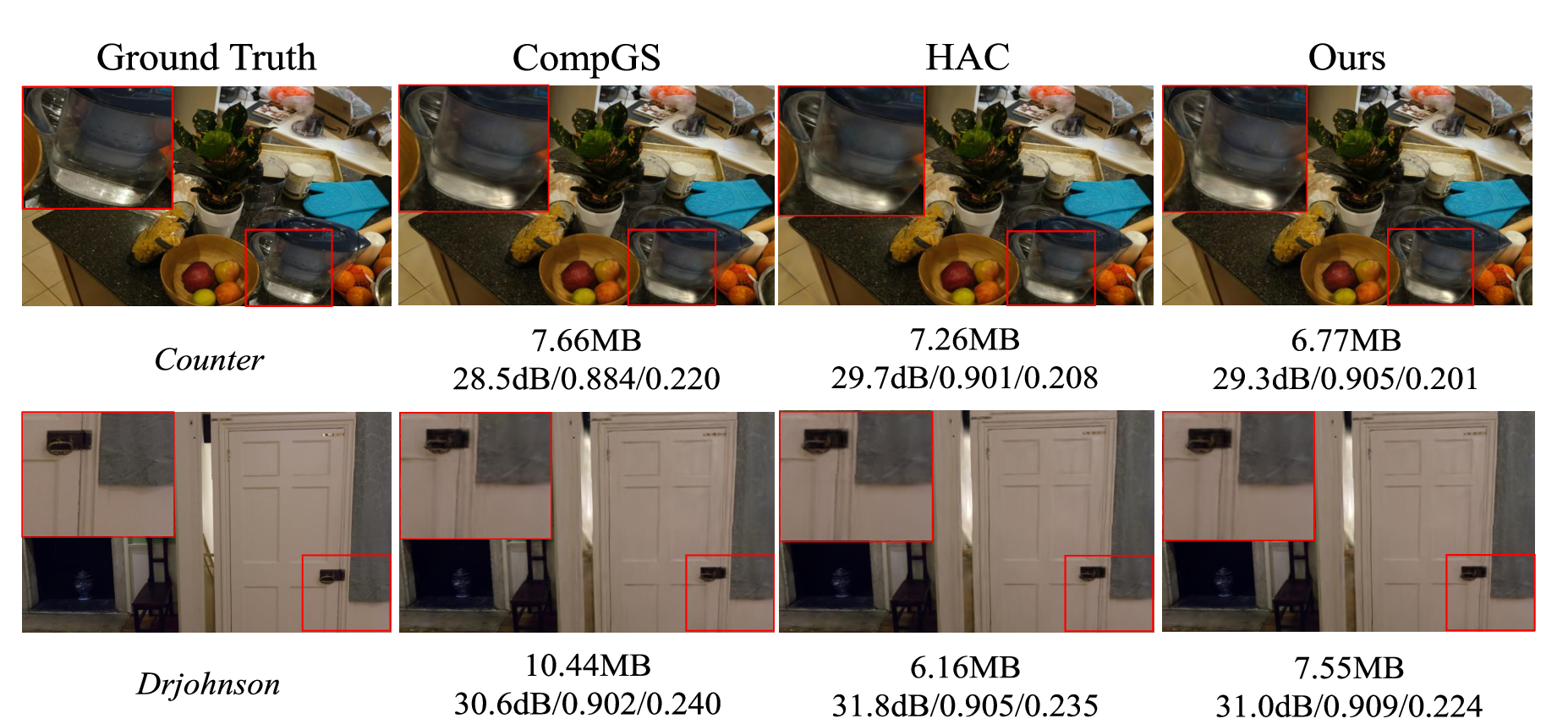}
 \caption{\textbf{Qualitative results for visual comparison.} Each subfigure displays the storage size along with the PSNR, SSIM, and LPIPS metrics.}
 \label{fig:fig6}
\end{figure*}
\section{Experiments and Results}
We first present the implementation details, followed by evaluation experiments to compare our method with existing 3DGS compression approaches. Additionally, we provide ablation studies to demonstrate the effectiveness of each technical component.

\subsection{Implementation Details}
\label{sec:sec4.1}

\textbf{Experimental Configurations.} After point initialization, we train the feature plane for 40k iterations. Each feature plane has 8 channels and a 512$\times$512 resolution with a decoder MLP to predict each attribute (color, scale, rotation, opacity) separately. Therefore, a total of 32 channels are used to predict all attributes. To prepare the feature plane for video codec input, its channels are concatenated into 32 frames, normalized to the range $[0,1]$ and then scaled to 16-bit integers. For progressive training, we use $T_i$ = \{0, 5000, 10000, 15000\} and $L_i$ = \{2, 4, 6, 8\}. Consequently, all channels are activated in training after 15k iterations. Entropy modeling, which involves DCT operations across all channels, can be computationally intensive. To mitigate computational intensity, we apply entropy loss only after the 30k iteration. We observed that performance converges properly with this condition. We also calculate channel importance score at 30k iteration and determine the weights $w_{c}$. We conducted all experiments with a single NVIDIA A100 GPU with 40GB memory.

\textbf{Feature Plane Compression.} For video coding, we use HEVC Test Model (HM) \cite{hevc_hm} version 16.0 with the random access (RA) configuration. The default RA configuration includes predefined QP offsets between frames to improve general video compression performance, but this may not be suitable for the feature plane compression. Therefore, we set all QP offsets between frames to 0. Since the feature plane is already highly optimized with entropy loss training, we expect minimal quality loss in video coding. Therefore, the QP is set to 1. In addition, we use the extension flag in HM to support the 16-bit image format, enabling the concatenated feature planes to be processed as a 16-bit grayscale (YUV400) format. The detailed configurations and commands are provided in the appendix.

\textbf{Point Position Compression.} Similar to the feature plane, the point positions are packed into a 512$\times$512 resolution, with the number of channels adjusted to accommodate the total number of points. Before packing, we apply Morton order sorting \cite{morton1966} to the positions, which are then quantized to 16-bit integers. Because point positions are critical for rendering quality, no regularization is applied and `lossless’ setting is used for coding.

\textbf{Piecewise-projective Contraction.} To map unbounded 360 scenes onto a plane, we utilized the piecewise-projective contraction introduced in \cite{merf10.1145/3592426}. Although originally proposed for handling ray-box intersections in grid-based structure, our experiments demonstrate that this contraction leads to a more compact representation of feature planes, by enhancing spatial correlation. Eq. (\ref{eq:eq1}) indicates a piece-wise projection, where $\|\mathbf{x}\|_{\infty}=\max _j\left|x_j\right|$.

\begin{equation}
\begin{aligned}
&\operatorname{contract}(\mathbf{x})_j= 
\begin{cases}
x_j & \text{if }\|\mathbf{x}\|_{\infty} \leq 1 \\[8pt]
\frac{x_j}{\|\mathbf{x}\|_{\infty}} & \text{if } x_j \neq\|\mathbf{x}\|_{\infty} > 1 \\[8pt]
\left(2 - \frac{1}{|x_j|}\right) \frac{x_j}{|x_j|} & \text{if } x_j = \|\mathbf{x}\|_{\infty} > 1
\end{cases}
\end{aligned}
\label{eq:eq1}
\end{equation}

\pagebreak

\textbf{Datasets.} We evaluate our method using the Mip-NeRF360 \cite{barron2022mip}, DeepBlending \cite{hedman2018deep}, and Tank\&Temples \cite{knapitsch2017tanks} datasets. These datasets comprise high-resolution multiview images captured from real-world scenes, featuring unbounded environments and complex objects. We follow the experimental conditions used in the original 3DGS.

\begin{figure}[t]
 \centering
      \begin{subfigure}[b]{0.49\linewidth}
         \centering
         \includegraphics[width=\linewidth]{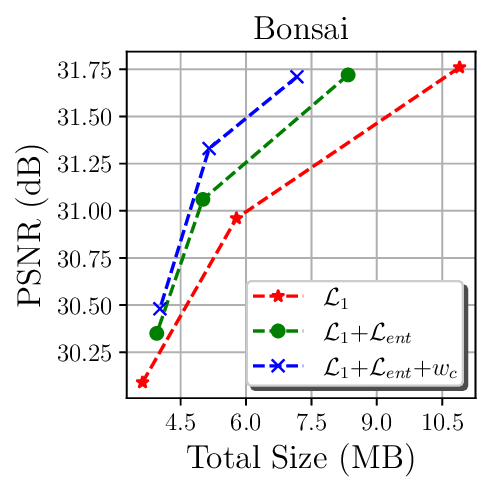}
         
     \end{subfigure}
     \vspace{0.01\linewidth}
          \begin{subfigure}[b]{0.49\linewidth}
         \centering
         \includegraphics[width=\linewidth]{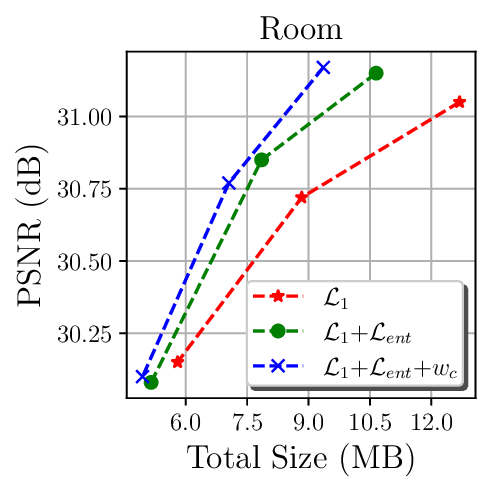}
         
     \end{subfigure}
     \vspace{0.01\linewidth}
          \begin{subfigure}[b]{0.49\linewidth}
         \centering
         \includegraphics[width=\linewidth]{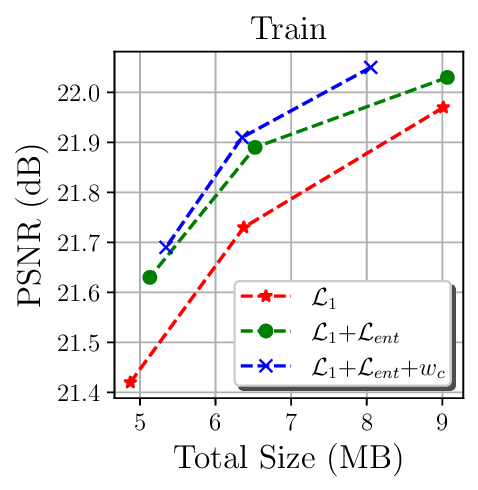}
         
     \end{subfigure}
     \vspace{0.01\linewidth}
          \begin{subfigure}[b]{0.49\linewidth}
         \centering
         \includegraphics[width=\linewidth]{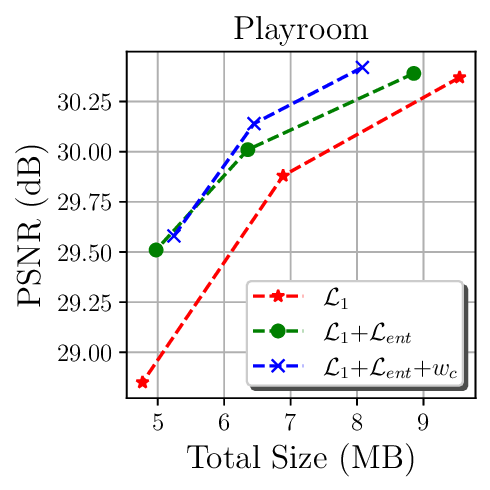}
     \end{subfigure}
         \caption{\textbf{RD curves from the ablation study demonstrate the effectiveness of our proposed method.} Based on $\mathcal{L}_{1}$, adding $\mathcal{L}_{ent}$ as a loss function provides a clear benefit. Moreover, introducing channel-wise bit allocation $w_{c}$ results in improved rate-distortion performance.
         }
 \label{fig:fig7}
\end{figure}

\subsection{Experimental Results}
\label{sec:sec4.2}
The proposed method is a 3DGS-based model that explicitly targets all gaussian attributes for compression. In contrast, HAC \cite{hac2024} and CompGS \cite{compgs10.1145/3664647.3681468} are based on Scaffold-GS \cite{lu2024scaffold} to explore the relationships between initialized gaussian parameters and predict offsets. As a result, these compression models exhibit different baseline quality levels compared to the original 3DGS. Tab. \ref{tab:tabs1} presents the quantitative comparison results of GS compression models. The proposed method demonstrates competitive performance even when compared to Scaffold-GS based compression models. Through a compact feature plane architecture, we achieved sizes under 10MB for all datasets, with no significant quality loss compared to the original 3DGS.

For the Mip-NeRF360 dataset, our method achieves compression ratios of up to 76$\times$ compared to 3DGS, with only 0.19dB loss in terms of PSNR. For the DeepBlending dataset, our method surpasses the quality of the original 3DGS, and other compression methods also show similar performance. While our method results in lower PSNR than HAC, it demonstrates better performance in SSIM, LPIPS metrics. For the Tank\&Temples dataset, our method achieves comparable visual quality in terms of PSNR, but shows lower results in SSIM and LPIPS metrics compared to the original 3DGS. HAC and CompGS also demonstrate lower metrics than our proposed method.

To provide a more comprehensive comparison results, we present rate-distortion (RD) curves for each dataset. Fig. \ref{fig:fig5} shows that our approach achieves significant size reduction while maintaining competitive rendering quality compared to other methods. On the Mip-NeRF 360 dataset, our method consistently achieves significant reductions in bitrate consumption. On the DeepBlending dataset, HAC demonstrates the highest rendering quality and overall best performance. Notably, our method also surpasses the rendering quality of original 3DGS, achieving similar results to other compression techniques. Our method outperforms CompGS at high bitrates. A comparable trend is observed in the Tank\&Temples dataset, where our approach demonstrates visual quality on par with HAC at low bitrates.

We also present qualitative results in Fig. \ref{fig:fig6}. By exploiting spatial correlations, our method effectively preserves texture details in a small size, resulting in a more compact representation of the compressed scene. Compared to CompGS, our method show better performance especially on the Mip-NeRF360 dataset. It is worth noting that our model has a slightly lower PSNR than HAC, but it shows competitive performance in other visual metrics. One possible explanation is that the performance of HAC and CompGS relies on the Scaffold-GS. As shown in Table \ref{tab:tabs1}, Scaffold-GS shows a higher LPIPS compared to the original 3DGS, particularly on the DeepBlending dataset. Contrary to HAC, our method easily integrates with video codecs, enabling more efficient real-world deployment and applications across various hardware.

\subsection{Ablation Study}
\label{sec:sec4.3}

\begin{table}[t]
\centering
  \centering
  \resizebox{0.45\textwidth}{!}{
\begin{tabular}{ccccc|cc}
\hline
PC          & $w_{c}$      & $\mathcal{L}_{ent}$ & PR         & $\mathcal{L}_1$ & PSNR(dB) & Size (MB)  \\ \hline
             &              &                     &              & $\checkmark$    & 27.27     & 23.45 \\
             &              &                     & $\checkmark$ & $\checkmark$    & 27.40     & 22.81 \\
             &              & $\checkmark$        & $\checkmark$ & $\checkmark$    & 27.31     & 10.68 \\
             & $\checkmark$ & $\checkmark$        & $\checkmark$ & $\checkmark$    &  27.29         &  9.96     \\
$\checkmark$ & $\checkmark$ & $\checkmark$        & $\checkmark$ & $\checkmark$    &  27.30         &  9.78    \\ \hline
\end{tabular}
}
\caption{\textbf{Ablation study on the proposed contributions for the Mip-NeRF360 dataset.} 'PR' denotes progressive training, while 'PC' represents piecewise-projective contraction as an alternative to unit sphere contraction \cite{barron2022mip}.}
\label{tab:tabs2}
\end{table}

{\textbf{Effectiveness of the $\mathcal{L}_{ent}$}. Since the feature plane is randomly initialized, unoccupied coordinates may have noisy values after training, potentially leading to undesirable results. We can incorporate $\mathcal{L}_{1}$ to encourage these areas to become zero. This property makes the loss function commonly used for achieving compact representations, resulting in smaller compression sizes \cite{tetrirf10656750}.

However, our experiments show that using only $\mathcal{L}_{1}$ for size adjustment results in significant quality degradation. We analyze how the $\mathcal{L}_{ent}$ term affects compression performance. This term encourages the transformed coefficient values to become a compact representation, which leads to a significant reduction in size. By incorporating our proposed $\mathcal{L}_{ent}$ with a fixed $\lambda_{1}$, we achieve significantly improved rate-distortion performance, as shown in Fig. \ref{fig:fig7}. 

Higher-level channels have a sparser representation, but this can cause significant differences between channels, making inter-prediction in video coding more difficult. With our method, the size of these channels can be reduced more efficiently. Further gains are achieved by introducing bit allocation to dynamically adjust $w_{c}$, thereby allocating fewer bits to less important channels.

\begin{table}
  \centering
  \resizebox{0.5\textwidth}{!}{
\begin{tabular}{c|c|c|cc}
\hline
Scene                     & Feature Plane & Positions    & Total Size      & PSNR           \\ \hline
\multirow{3}{*}{Kitchen}  & 8.91   & 2.34 & 11.2               & 30.88               \\
                          & 6.73    & 2.34  & 9.07 & 30.56 \\
                          & 3.64    & 2.34  & 5.98 & 29.96 \\ \hline
\multirow{3}{*}{Playroom} & 9.31     & 2.87  & 12.1                & 30.38              \\
                          & 6.45   & 2.87  &  9.32                & 30.17                \\
                          & 3.62   & 2.87 &  6.49                 & 28.92                \\ \hline
\multirow{3}{*}{Truck}    & 8.81   & 3.35  & 12.2                 & 25.19               \\
                          & 6.25   & 3.35 & 9.60                 & 25.05               \\
                          & 4.58    & 3.35  & 7.93                 & 24.28                \\ \hline
\end{tabular}
}
\caption{\textbf{Ablation studies on the size proportions of feature planes and point positions for scenes from each dataset.} We use $\lambda_{ent}$ to adjust the trade-off between the compressed size of the feature plane and visual quality.}
\label{tab:tabs3}
\end{table}

\textbf{Evaluation of the Proposed Components.} Tab. \ref{tab:tabs2} presents an ablation study on the proposed components of the method for the Mip-NeRF360 dataset. For better comparison clarity, we adjusted the hyperparameters to achieve results with similar visual quality. This demonstrates that integrating each component effectively improves both compression and rendering quality, with the resulting configuration producing a more compact representation.

{\textbf{Feature Plane Size Proportion.} Tab. \ref{tab:tabs3} shows the results for the size proportions of feature planes and point positions at different compression parameters. The proposed method does not affect the densification process, so the size of the point positions remains consistent across scenes. As we compress to lower bitrates with higher $\mathcal{L}_{ent}$ values, the proportion of the point position size gradually increases. Because the proposed method does not involve pruning, the number of points remains the same as in the original 3DGS. However, because our model is independent of the densification process, there is potential to reduce position size by adjusting the number of points. 

{\textbf{Video Codec Performance.} While this paper presents experimental results using HM, it is important to note that HM is not optimized for real-time performance. Tab. \ref{tab:tabs4} compares the results with FFmpeg libx265 \cite{ffmpeg}, which supports hardware-accelerated encoding and decoding, significantly speeding up video coding for HEVC features. As shown, the proposed method consistently improves performance across the two codec implementations. Since our method effectively leverages any DCT-based standard codec technology, there is potential for further improvements as video codecs continue to advance. We also observe that coding complexity is slightly reduced due to the more compact representation. 

\textbf{Complexity Overhead.} The main limitation of the proposed method is its increased training time, which is based on the slow convergence speed of the grid-based approach. For the Mip-NeRF360 dataset, including the point initialization stage, training takes about 90 minutes per scene. However, once all attributes are predicted from the decompressed feature plane, no additional overhead is required for rendering. Since the proposed method follows the same densification process as the original 3DGS, its rendering time is comparable to that of the original model.


\begin{table}[t]
  \centering
  \resizebox{0.5\textwidth}{!}{
\begin{tabular}{c|cc|cl}
\hline
Video codec              & Size (MB) & PSNR (dB) & \multicolumn{2}{c}{Enc/Dec time (s)} \\ \hline
HM \cite{hevc_hm} (w/o ours)     & 27.9      & 23.71     & \multicolumn{2}{c}{292.9/2.9}       \\
HM \cite{hevc_hm} (ours)          & 7.46      & 23.63     & \multicolumn{2}{c}{267.8/2.8}       \\ \hline
libx265 \cite{ffmpeg} (w/o ours) & 29.2      & 23.72     & \multicolumn{2}{c}{27.5/0.7}         \\
libx265 \cite{ffmpeg} (ours)     & 7.79      & 23.59     & \multicolumn{2}{c}{25.3/0.6}        \\ \hline
\end{tabular}
}
\caption{\textbf{Comparison results for different codecs on the Tank\&Temples Dataset.} The proposed method shows consistent performance improvements with conventional video codecs.}
\label{tab:tabs4}
\end{table}

\section{Conclusion}
\label{sec:conclusion}
We present a compression format in this paper that uses standard video codecs to efficiently handle 3D gaussian attributes for compact scene representations. By combining feature planes with entropy modeling in the frequency domain, our approach achieves significant size reduction without compromising rendering quality. The proposed entropy loss function works effectively with video codecs, enabling the reuse of advanced coding techniques. This feature plane-based compression format makes a wide range of applications possible on mobile devices and other platforms with codec support. As a result, our method provides key insights that can drive the development of 3D representations using video codecs, leading to more efficient 3D compression. Experimental results validate our method, demonstrating substantial storage savings while maintaining high visual fidelity.
\clearpage
{
    \small
    \bibliographystyle{ieeenat_fullname}
    \bibliography{main}
}
\clearpage

\clearpage
\maketitlesupplementary

\textbf{Detailed Architecture.} Each attribute is decoded using an MLP $g$, a three-layer, fully connected network. The intermediate layers contain 128 units each and use ReLU activation, except for the output layer. The four decoder MLPs have a total size of 0.28 MB, which is included in the final size results. We applied a sigmoid function to ensure the opacity values are within the range $[0, 1]$. For scaling, we first used the sigmoid function, then scaled the values to the range $[-10, -0.1]$, and followed with an exponential function. Finally, the predicted attributes are used as inputs to the original 3DGS rasterizer.

The feature plane learning rate is set to 0.005 using the Adam optimizer. Additionally, we apply learning rate schedulers to gradually decrease the learning rates during training. All other settings, including the learning rate for point positions, follow the original 3DGS.

\begin{table}[h]
\centering
\begin{tabular}{c|cc}
\hline
\text {$Q_{step}$} & Feature Plane Size (MB) & PSNR  \\ \hline
$2^{0}$ & 16.6 & 31.55 \\
$2^{2}$ & 14.2 & 31.66\\
$2^{4}$   & 11.5      & 31.54 \\
$2^{6}$    & 9.07      & 31.75 \\
$2^{8}$    & \textbf{8.45}      & \textbf{31.72} \\
$2^{10}$     &   12.1         & 31.77     \\
$2^{12}$    &      17.8      & 31.32      \\ \hline
\end{tabular}
\caption{Effect of $Q_{step}$ on the proposed method for the 'bonsai' scene, with fixed $\lambda_{ent}$ across all experiments.}
\label{tab:stabs1}
\end{table}

\textbf{Analysis for Quantization Step Size.} To effectively optimize the standard video codec process, the quantization step size $Q_{step}$ is crucial. During training, the uniform quantizer with a $Q_{step}$ operates on floating-point transformed coefficients. However, it is important to note that the plane is scaled to 16-bit integers for compression. The difference between the pixel domain and the frequency domain makes it difficult to determine the proper quantization step size.  Tab. \ref{tab:stabs1} shows the impact of different quantization step sizes $Q_{step}$ on our proposed method's performance. Throughout all experiments, we maintained a constant $\lambda_{ent}$ to ensure consistent comparison. Since video codecs use predefined quantization matrix for compression, if the $Q_{step}$ is too small or too large, entropy modeling may not work correctly. Our experiments demonstrate that a $Q_{step}$ of $2^{8}$ produces better results.

\textbf{Configurations for Video Coding.} The configuration parameters of traditional codecs influence the performance of the model. We provide the detailed settings used to obtain our experimental results, ensuring reproducibility of our findings. 

The FFmpeg x265 command lines for lossless encoding of point positions used in our paper are:

\renewcommand{\ttdefault}{pcr}
\begin{lstlisting}[language=Bash]
ffmpeg
-y
-pix_fmt gray16be
-s {width}x{height}
-framerate {framerate}
-i {input file name}
-c:v libx265
-x265-params
lossless=1
{bitstream file name}
\end{lstlisting}

The command lines for HM 16.0 (RExt) compression are:

\begin{lstlisting}[language=Bash]
TAppEncoder
-c encoder_randomaccess_main_rext.cfg
--InputFile={input file name}
--SourceWidth={width}
--SourceHeight={height}
--InputBitDepth=16
--InternalBitDepth=16
--OutputBitDepth=16
--InputChromaFormat=400
--FrameRate={framerate}
--FramesToBeEncoded=32
--QP={qp}
--BitstreamFile={bitstream file name}
\end{lstlisting}

As mentioned in the manuscript, the HM configuration used in the experiments has all QP offsets between frames set to 0. All other settings followed the default configuration. The modified configuration is as follows:

\begin{lstlisting}[language=Bash, mathescape=true]
encoder_randomaccess_main_rext.cfg
#        Type POC QPoffset (...)
Frame1:  B   16   $\textbf{0}$
Frame2:  B    8   $\textbf{0}$
Frame3:  B    4   $\textbf{0}$
Frame4:  B    2   $\textbf{0}$
Frame5:  B    1   $\textbf{0}$
Frame6:  B    3   $\textbf{0}$
Frame7:  B    6   $\textbf{0}$
Frame8:  B    5   $\textbf{0}$
(...)
\end{lstlisting}

\begin{figure*}[t]
 \centering
      \begin{subfigure}[b]{0.24\linewidth}
         \centering
         \includegraphics[width=0.98\linewidth]{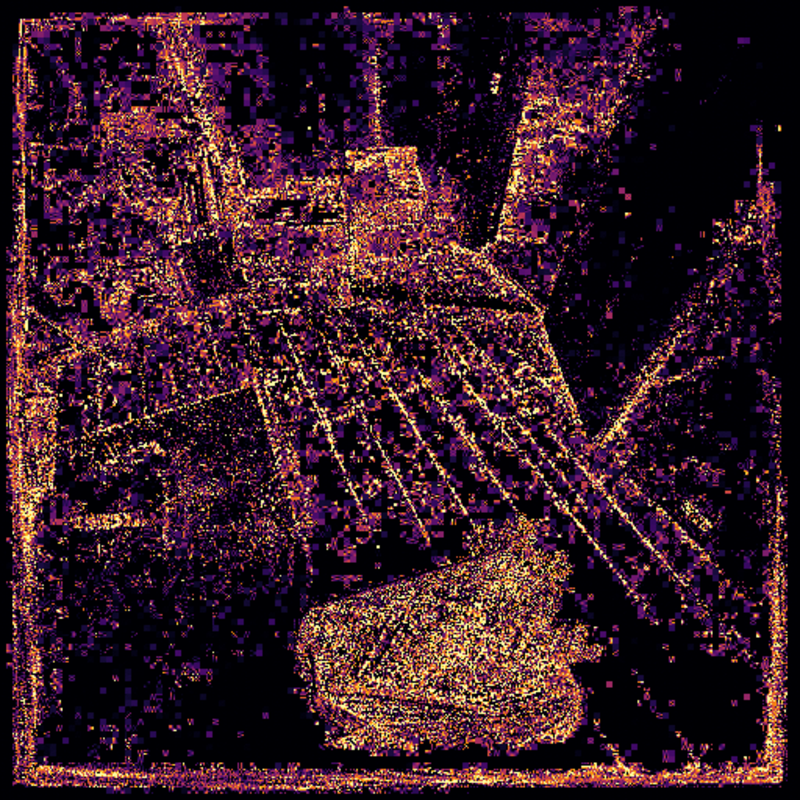}
         \caption{$\mathcal{P}_{1}^{XZ}$}
     \end{subfigure}
     \vspace{0.01\linewidth}
          \begin{subfigure}[b]{0.24\linewidth}
         \centering
         \includegraphics[width=0.98\linewidth]{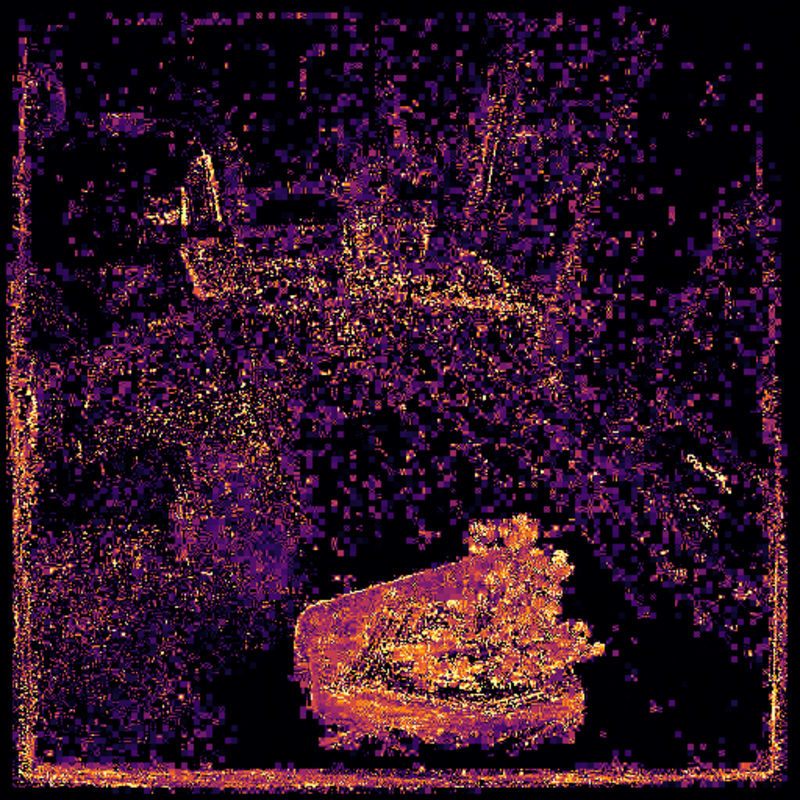}
         \caption{$\mathcal{P}_{3}^{XZ}$}
     \end{subfigure}
     \vspace{0.01\linewidth}
          \begin{subfigure}[b]{0.24\linewidth}
         \centering
         \includegraphics[width=0.98\linewidth]{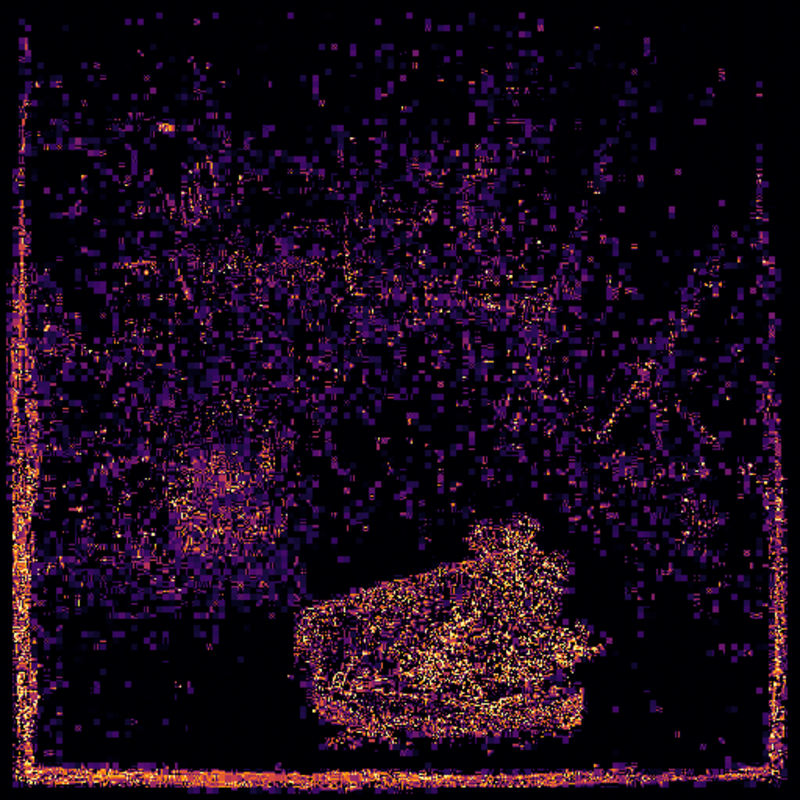}
         \caption{$\mathcal{P}_{5}^{XZ}$}
     \end{subfigure}
     \vspace{0.01\linewidth}
          \begin{subfigure}[b]{0.24\linewidth}
         \centering
         \includegraphics[width=0.98\linewidth]{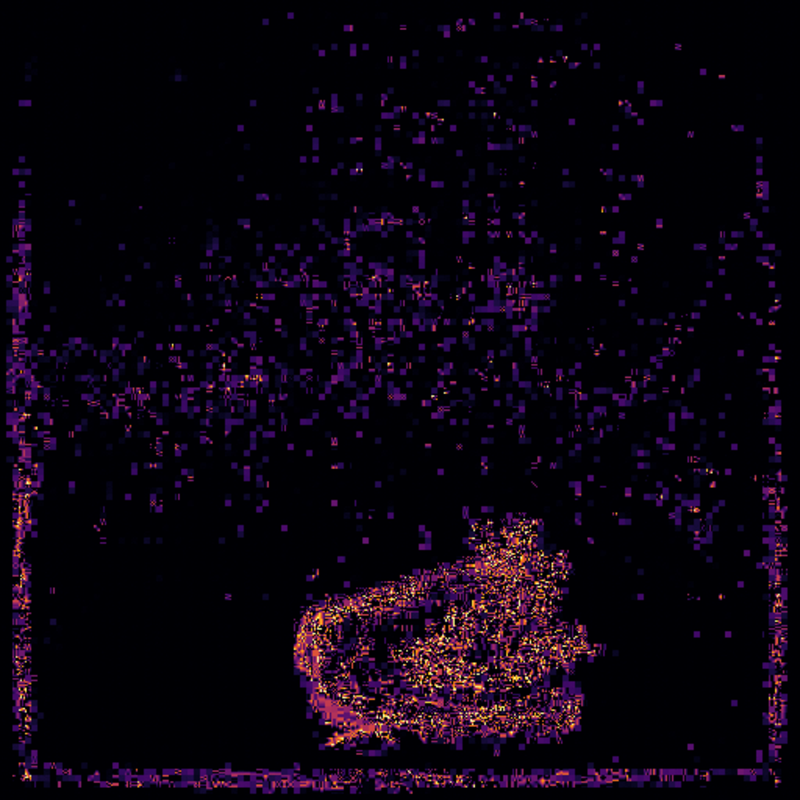}
         \caption{$\mathcal{P}_{7}^{XZ}$}
     \end{subfigure}

 \caption{\textbf{Visualization of channel levels in XZ plane for the \textit{`Bonsai'} scene.} With dynamic $w_{c}$, the lower-level channels preserve more information, whereas the higher-level channels are largely minimized due to the higher $\lambda_{ent}$ weight assignment.
 }
 \label{fig:sfig1}
\end{figure*}

\pagebreak

\textbf{Analysis of Bit Allocation.} Fig. \ref{fig:sfig1} illustrates the results of each channel learned through our proposed bit allocation method. The signals in higher-level channels exhibit increased sparsity, making them more challenging for video coding. However, since these channels have lower impact on visual quality, most regions are minimized through allocating higher entropy $\lambda_{ent}$ for each level. This strategic allocation results in improved rate-distortion performance compared to conventional approaches.

\begin{figure}[h]
 \centering
      \begin{subfigure}[b]{\linewidth}
         \centering
         \includegraphics[width=\linewidth]{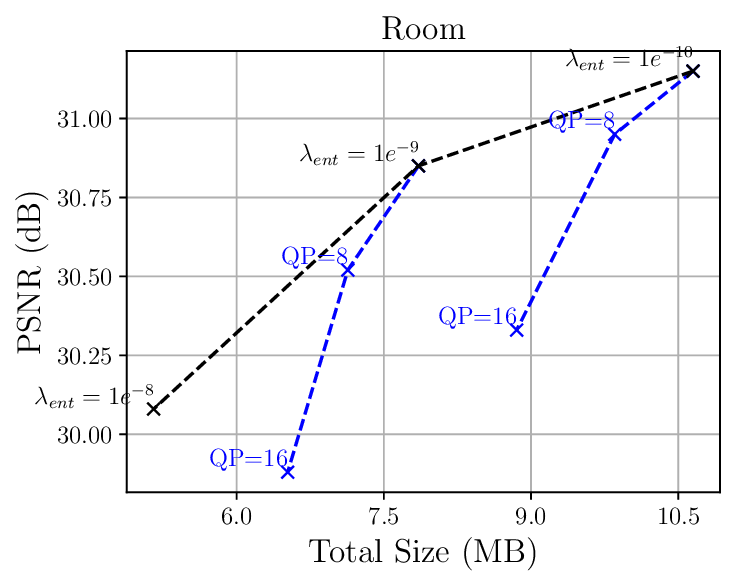}
         
     \end{subfigure}
         \caption{\textbf{RD curves with video QP adjustment.} Controlling rate-distortion with video QP is worse than using $\lambda_{ent}$ with QP=1.
         }
\label{sfig2}
\end{figure}

\textbf{Performance analysis of QP adjustment.} Beyond adjusting the parameter $\lambda_{ent}$, the rate-distortion trade-off can also be controlled by modifying the quantization parameter (QP) in video codecs such as HM or FFmpeg. However, our experimental results demonstrate that modifying the video codec QP yields inferior performance compared to adjusting $\lambda_{ent}$, as shown in Fig. \ref{sfig2}. Our results suggest this may be because video codecs focus exclusively on feature plane restoration rather than the rendering view quality.

\begin{figure}[h]
 \centering
      \begin{subfigure}[b]{0.49\linewidth}
         \centering
         \includegraphics[width=0.98\linewidth]{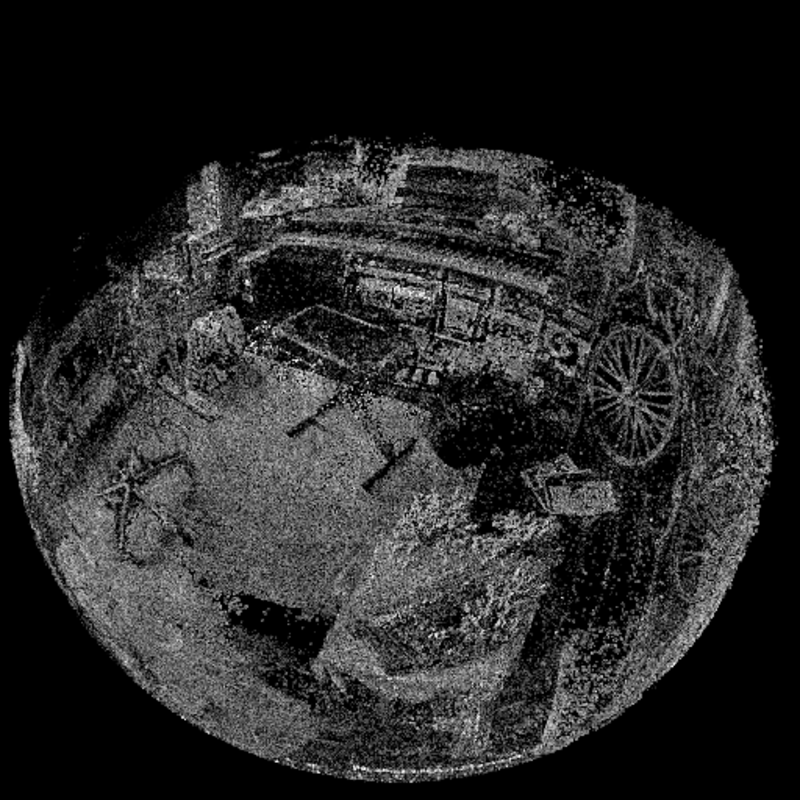}
         \caption{sphere contraction}
     \end{subfigure}
     \vspace{0.01\linewidth}
          \begin{subfigure}[b]{0.49\linewidth}
         \centering
         \includegraphics[width=0.98\linewidth]{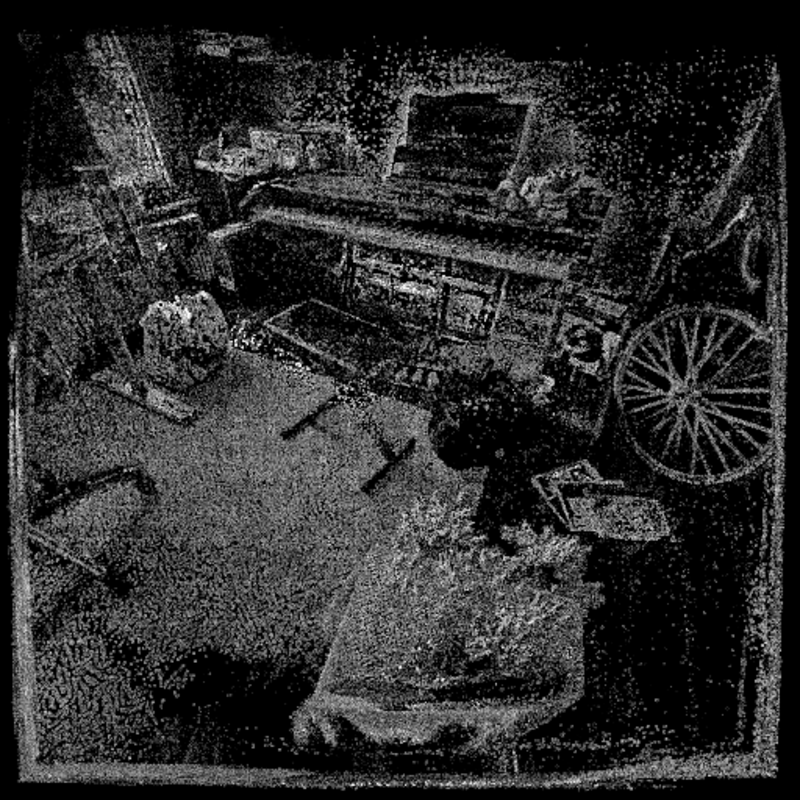}
         \caption{piecewise contraction}
     \end{subfigure}
 \caption{\textbf{Visualization of each contraction method for the \textit{`Bonsai'} scene.} (a) sphere contraction and (b) piecewise-projective contraction.}
 
\label{sfig3}
\end{figure}

\textbf{Contraction methods.} Fig \ref{sfig3} presents the feature plane results for each contraction strategy. Although there is no significant difference in reconstruction quality between the two methods, piecewise contraction is expected to perform better in block-wise DCT due to the spatial correlation of the features. Experimental results demonstrate a slight improvement in size reduction when using piecewise contraction.

\textbf{Per-scene Quantitative Results.} We evaluated the performance on various datasets for novel view synthesis. Our analysis includes per-scene results for the Mip-NeRF 360, Deep Blending, and Tank\&Temples datasets.

\begin{table*}[h]
\centering
\begin{tabular}{c|cccccccccc}
\hline
Scene        & bicycle & flowers & garden & stump & tree hill & room  & counter & kitchen & bonsai & Avg. \\ \hline
PSNR         & 25.19   & 21.37   & 27.61  & 26.59 & 23.07     & 30.99 & 28.47   & 30.73   & 31.71  & 27.30         \\
SSIM         & 0.746   & 0.593   & 0.853  & 0.781 & 0.648     & 0.922 & 0.898   & 0.916   & 0.934  & 0.810         \\
LPIPS        & 0.265   & 0.380   & 0.133  & 0.236 & 0.346     & 0.217 & 0.208   & 0.137   & 0.202  & 0.236         \\
Storage (MB) & 9.82    & 10.25   & 15.22  & 14.30 & 9.92      & 7.91  & 6.25    & 7.68    & 6.71   & 9.78          \\ \hline
\end{tabular}
\caption{Per-scene results evaluated on Mip-NeRF 360 dataset.}
\end{table*}

\begin{figure*}[h]
 \centering
      \begin{subfigure}[b]{\linewidth}
         \centering
         \includegraphics[width=\linewidth]{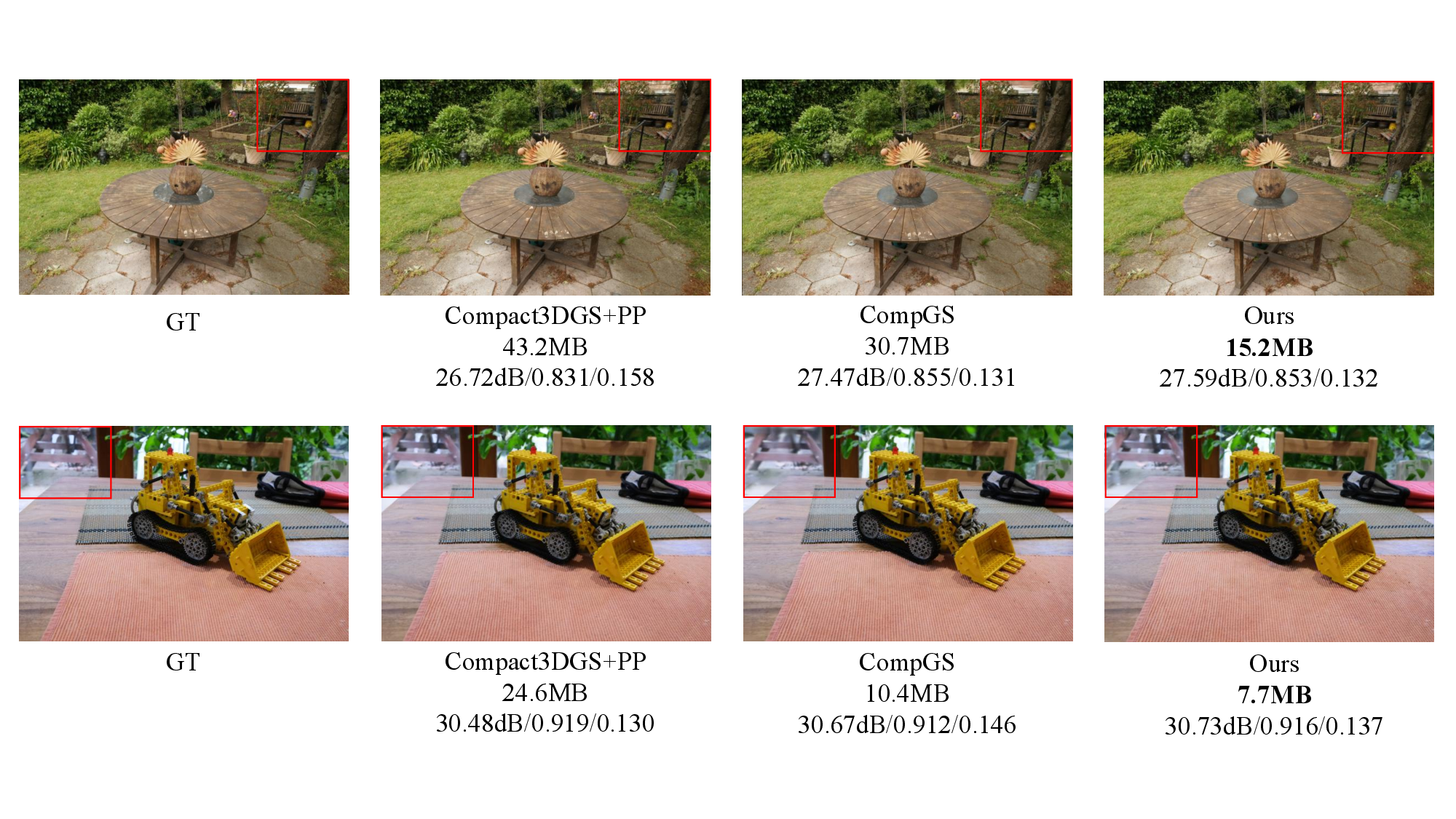}
         
     \end{subfigure}
         \caption{\textbf{Qualitative results for visual comparison for Mip-NeRF360 dataset.} Each subfigure displays the storage size along with the PSNR, SSIM, and LPIPS metrics. Detailed observation is encouraged by zooming in.
         }
\label{sfig3}
\end{figure*}

\clearpage
\pagebreak

\begin{table*}[h]
\centering
\begin{tabular}{c|cccccc}
\hline
Dataset      & \multicolumn{3}{c}{Tanks\&Temples}   & \multicolumn{3}{c}{Deep Blending}           \\ \hline
Scene        & train & truck & Avg.        & drjohnson & playroom & Avg.        \\ \hline
PSNR         & 22.08 & 25.19 & 23.63                & 29.22     & 30.42    & 29.82              \\
SSIM         & 0.801 & 0.882 & 0.842 & 0.904     & 0.909    & 0.907 \\
LPIPS        & 0.226 & 0.158 & 0.192 & 0.250     & 0.252   & 0.251 \\
Storage (MB) & 6.21  & 8.72  & 7.46                 & 9.39      & 7.86     & 8.62                 \\ \hline
\end{tabular}
\caption{Per-scene results evaluated on Tank\&Temples and Deep Blending.}
\end{table*}

\begin{figure*}[t]
 \centering
      \begin{subfigure}[b]{\linewidth}
         \centering
         \includegraphics[width=\linewidth]{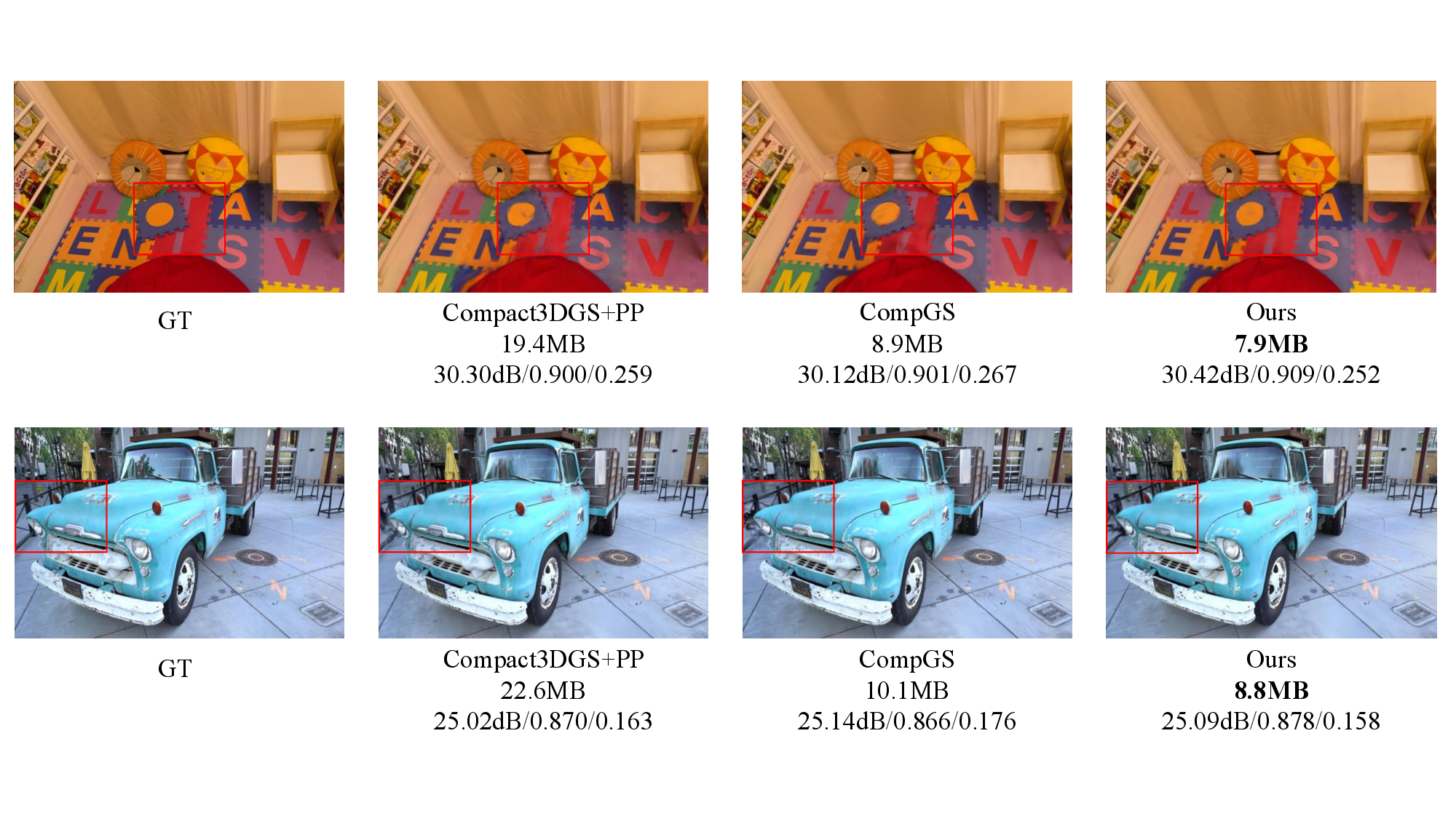}
         
     \end{subfigure}
         \caption{\textbf{Qualitative results for visual comparison for DeepBlending and T\&T dataset.} Each subfigure displays the storage size along with the PSNR, SSIM, and LPIPS metrics. Detailed observation is encouraged by zooming in.
         }
\label{sfig4}
\end{figure*}

\end{document}